\documentclass[10pt,twocolumn,letterpaper]{article}

\usepackage{cvpr}
\usepackage{times}
\usepackage{epsfig}
\usepackage{graphicx}
\usepackage{amsmath}
\usepackage{amssymb}
\usepackage{booktabs}
\usepackage{array}
\usepackage{tablefootnote}
\usepackage{float}
\usepackage{cuted}
\usepackage{multirow}
\usepackage{adjustbox}
\usepackage{flushend}
\usepackage[font=small,labelfont=bf,tableposition=top]{caption}
\DeclareMathOperator*{\argmax}{arg\,max}

\usepackage[pagebackref=true,breaklinks=true,letterpaper=true,colorlinks,bookmarks=false]{hyperref}

\cvprfinalcopy %

\def\cvprPaperID{2752} %

\ifcvprfinal\fi

\newcommand\minisection[1]{\vspace{2mm}\noindent \textbf{#1}}

\newcommand{\modelname}{HD$^3$\xspace}
\newcommand{\flowmodelname}{HD$^3$F\xspace}

\newcommand{\stereomodelname}{HD$^3$S\xspace}

\newcommand{\vtodname}{$\mathcal{V}2\mathcal{D}$\xspace}
\newcommand{\dtovname}{$\mathcal{D}2\mathcal{V}$\xspace}
\newcommand{\plres}{$p^l_{\subfix{res}}$\xspace}

\newcommand\vf[1]{\mathbf{#1}}

\newcommand\subfix[1]{\mathtt{#1}}

\begin{document}

\title{Hierarchical Discrete Distribution Decomposition for Match Density Estimation}

\author{Zhichao Yin \quad Trevor Darrell \quad Fisher Yu \quad \vspace{1.5mm} \\
UC Berkeley}

\maketitle

\begin{abstract}
Explicit representations of the global match distributions of pixel-wise correspondences between pairs of images are desirable for uncertainty estimation and downstream applications. However, the computation of the match density for each pixel may be prohibitively expensive due to the large number of candidates.
In this paper, we propose Hierarchical Discrete Distribution Decomposition (\modelname), a framework suitable for learning probabilistic pixel correspondences in both optical flow and stereo matching. We decompose the full match density into multiple scales hierarchically, and estimate the local matching distributions at each scale conditioned on the matching and warping at coarser scales. The local distributions can then be composed together to form the global match density. Despite its simplicity, our probabilistic method achieves state-of-the-art results for both optical flow and stereo matching on established benchmarks. We also find the estimated uncertainty is a good indication of the reliability of the predicted correspondences.

\end{abstract}
\vspace{-4ex}

\section{Introduction}
\begin{figure*}[t]
\begin{center}
   \includegraphics[clip, trim=0cm 0cm 0cm 0cm, width=1.0\textwidth]{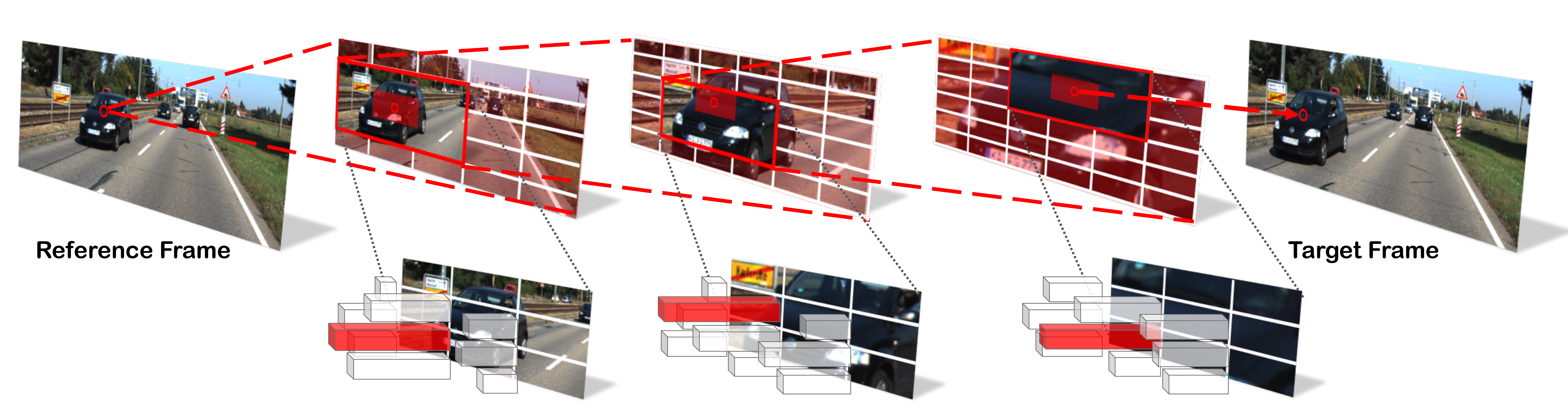}
\end{center}
\vspace{-1ex}
   \caption{Illustration of \modelname. We aim to estimate discrete match distribution in this work. For reducing the infeasible computation cost, the overall distribution is decomposed into multiple scales hierarchically at learning time.
The full match information can be recovered by composing predictions from all levels. Please refer to Sec.~\ref{sec::hdd} for more details.} %
   \vspace{-2ex}
\label{fig::teaser}
\end{figure*}

Finding dense pixel correspondences between two images, typically for stereo matching or optical flow, is one of the earliest problems studied in the computer vision literature. Dense correspondences have  wide application including for activity recognition~\cite{sevilla2018integration}, video interpolation~\cite{jiang2018super},  scene geometry perception~\cite{geiger2013vision}, and many others. Challenges when solving this problem include texture ambiguity, complex object motion, illumination change, and entangled occlusion estimation.

Classic approaches jointly optimize local texture matching and neighbor affinity on  images~\cite{horn1981determining} possibly in a coarse-to-fine fashion~\cite{behl2017bounding,mirror,prsm}. While these methods can achieve impressive correspondence accuracy, the optimization step may be too slow for downstream applications. Recent works using deep convolutional networks (ConvNets) have achieved similar or even better matching results without an optimization step~\cite{dosovitskiy2015flownet,ilg2017flownet,sun2018pwc}. Pixel features learned directly from correspondence supervision can capture both local appearance and global context information due to the large network receptive fields. With GPU acceleration, it is possible to use these networks to regress the pixel displacements in real time~\cite{dosovitskiy2015flownet,sun2018pwc}.

However, the estimation uncertainty inherent in correspondence estimation is neglected by  displacement regression approaches. 
Though post-hoc confidence measures~\cite{kondermann2008statistical,mac2013learning} can recover the uncertainty to some degree, they are independent of model training;  uncertainty is  ignored in the training process. Recognizing the missing uncertainty measures in optical flow methods, some works~\cite{gast2018lightweight,wannenwetsch2017probflow} propose probabilistic frameworks for joint correspondence and uncertainty estimation. Due to constraints on computation and parameter number, they rely on the local Gaussian noise assumption to represent the match distribution. Consequently, they cannot model complicated distributions on a large image area. Early works in stereo matching show that we can build a complete match cost volume as a proxy to estimate the match density, but were not applicable for high-resolution stereo matching nor general optical flow due to the excessive amount of computation needed for the complete cost volume.

In this work, we propose Hierarchical Discrete Distribution Decomposition (\modelname), a general probabilistic framework for match density estimation. We aim to find the discrete distribution of possible correspondences with a large support defined on the image grids for each pixel. We adopt a general model to represent  pixel-level match probability without any parametric distribution assumption. The model-inherent uncertainty measures can be naturally derived from our estimated match densities.

\modelname decomposes the full match density into multiple levels of local distributions similar to quadtrees. To extract discriminative features for matching, we use networks with Deep Layer Aggregation (DLA)~\cite{yu2018deep} to build the multi-scale feature pyramid. The DLA framework provides us with feature networks of different computation-accuracy trade-offs, which can be easily integrated with other recognition tasks in complex applications.
We estimate the match density of the residual motion in each scale, conditioned on match densities at coarser scales. We can propagate the conditional information from previous levels to the prediction at the current level through iterative feature warping and density bypass connections. The multi-scale match densities can then be used to recover the complete match density.
We can easily convert between point estimates and match densities to train our models on existing datasets with annotations in the form of motion vectors.

We evaluate our framework extensively in two 
applications: stereo 
matching and optical flow. 
Our method achieves state-of-the-art results on both the synthetic dataset MPI Sintel~\cite{sintel} and the real dataset KITTI~\cite{geiger2013vision}. Our method not only surpasses all two-frame based optical flow methods by large margins but also beats some competitive scene flow methods on both KITTI 2012 \& 2015. We also evaluate our uncertainty estimation and demonstrate the error-awareness of our method in its predictions. Our code is available at  \url{https://github.com/ucbdrive/hd3}.

\section{Related Work}
\begin{figure*}[t]
\begin{center}
   \includegraphics[clip, trim=0cm 0cm 0cm 0cm, width=0.8\textwidth]{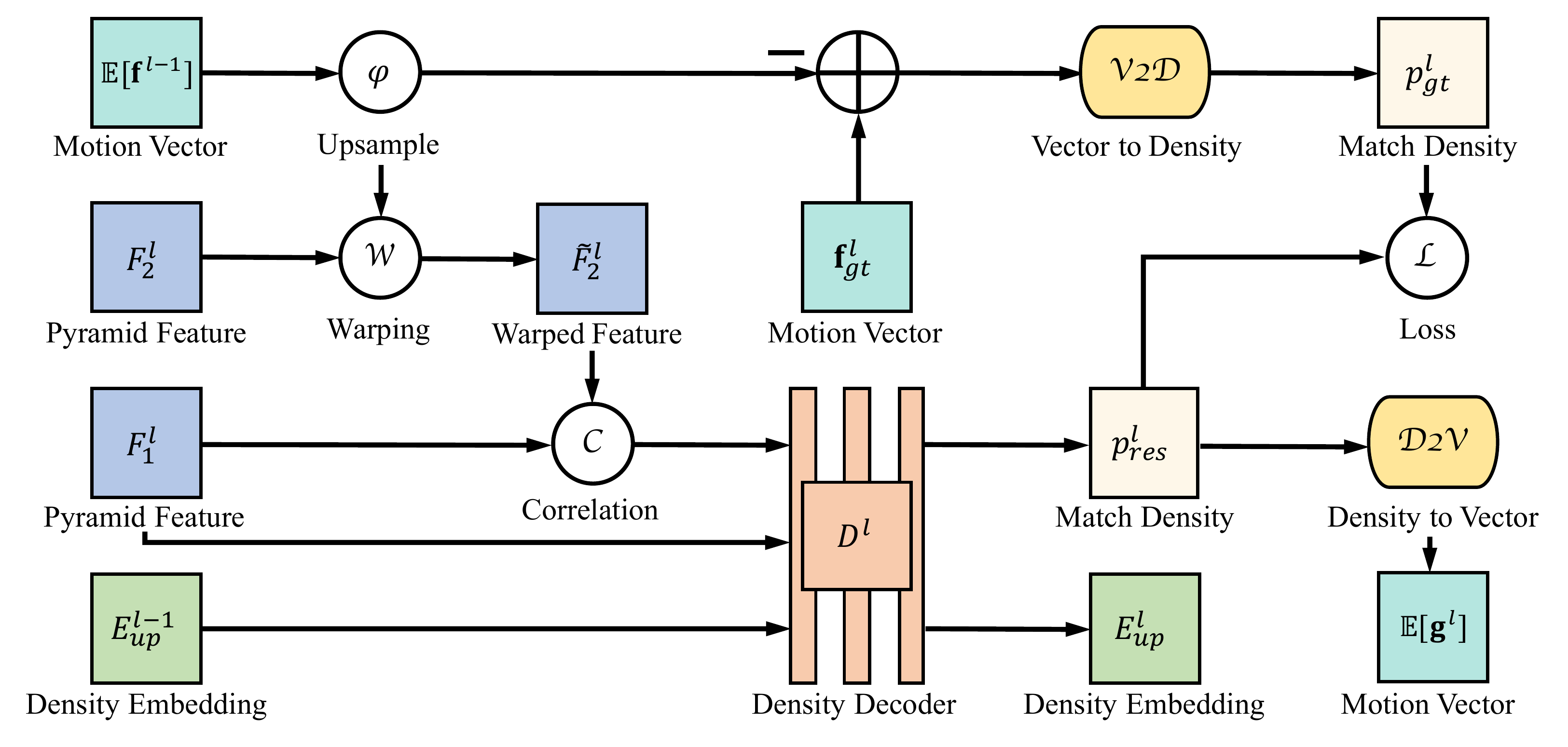}
\end{center}
\vspace{-2ex}
   \caption{Overview of our architecture. The submodule at the $l_{\subfix{th}}$ level is presented here. $F^l$ and $\tilde{F}^l$ denotes the $l_{\subfix{th}}$ level of original and warped pyramid features of image pair $I$. $E_{\subfix{up}}^l$ denotes upsampled density embeddings between different levels as density bypass connections. $\vf{f}^l$ and $\vf{g}^l$ denote motion vectors and $p^l$ corresponds to match density. Their conversion is fulfilled by our \dtovname and \vtodname modules. %
For details please refer to our method part. This figure is best viewed in color.}
   \vspace{-2ex}
\label{fig::pipeline}
\end{figure*}

Great efforts have been devoted to the problems of finding dense correspondences in the past four decades. 
For a thorough review, we refer to popular benchmarks including Middlebury~\cite{baker2011database}, MPI Sintel~\cite{sintel}, and KITTI~\cite{menze2015object} benchmarks for both the classical methods and the latest advances in these areas. We will discuss the most related ideas in this section.

\minisection{Correspondence Estimation.} 
Classical stereo matching usually involves local correspondence extraction and semi-global regularization~\cite{sgm}. 
On the other hand, optical flow methods typically adopt MRFs~\cite{li1994markov} to jointly reason about the displacements, occlusions, and symmetries~\cite{mirror,sps} for tackling the more unconstrained and challenging 2D correspondence problem. Despite the distinct differences between the search space dimensions, stereo matching and optical flow share similar assumptions such as brightness constancy and edge-preserving continuity~\cite{horn1981determining,epic,sun2018pwc}.

With the success of deep learning, end-to-end models have been designed for these dense prediction tasks. Benefiting from pretraining on a large corpus of synthetic data~\cite{dispnet}, these methods achieve impressive results on par with classical methods~\cite{dosovitskiy2015flownet,ilg2017flownet}. 
Furthermore, recent advances emphasize the incorporation of classical principles into network designs, such as pyramid matching, feature warping, and contextual regularizer~\cite{hui2018liteflownet,gcnet,sun2018pwc}. These improvements contribute to the superior performance of deep learning models, allowing them to surpass classical methods. 
However, such learning methods neglect the model-inherent uncertainty estimation,~\ie they are agnostic of the prediction failure, which is quite important for applications such as autonomous driving and medical imaging. 
In contrast, our work focuses on the probabilistic correspondence estimation, which can naturally convey the confidence of the predictions.

\minisection{Uncertainty Measures.}
Various uncertainty measures have been proposed for classical optical flow estimation. Barron~\etal~\cite{barron1994performance} proposed a simple method based on the input data characteristics while ignoring the estimated optical flow itself. Kondermann~\etal~\cite{kondermann2008statistical} learned a probabilistic flow model and obtained uncertainty estimation through hypothesis testing. 
Mac Aodha~\etal~\cite{mac2013learning} trained a classifier to assess the prediction quality in terms of end-point-error. These methods either leverage only part of the input information, such as images or predicted flow, for uncertainty estimation, or require post-processing steps independent of model inference itself.

Recently, Gast~\etal~\cite{gast2018lightweight} recognized the importance of model-inherent uncertainty measures for deep networks. They proposed probabilistic output layers and employed assumed density filtering to propagate activation uncertainties through the network. For computational tractability, they assumed Gaussian noise and adopted a parametric distribution. Though it can be easily adapted for use with existing regression networks, their performance is only competitive with the deterministic counterparts. 
Our method provides inherent uncertainty estimation as well as new state-of-the-art results for both optical flow and stereo matching.

\minisection{Coarse-to-Fine.} 
Because of the complexity of finding 2D correspondences for each pixel in optical flow, it is natural to match the pixels from coarse to fine resolutions in an image or feature pyramid. 
This method can be used effectively in the optimization methods~\cite{anandan1989computational,bailer2015flow,simoncelli1991probability} as well as patch matching~\cite{cpm}. 
Its effectiveness is also verified by recent deep learning approaches such as SpyNet~\cite{ranjan2017optical}, PWC-Net~\cite{sun2018pwc}, and LiteFlowNet~\cite{hui2018liteflownet}. 
We also estimate our hierarchically decomposed match densities based on the feature pyramid representation~\cite{yu2018deep}. 
Our contribution lies in the decomposition of the discrete probability distribution instead of the feature representations.

\section{Method}
In this section, we discuss 
our probabilistic framework for match density estimation. 
Without loss of generality, we focus on solving 2D correspondences for optical flow in this section, which can be easily adapted to the 1D case for stereo matching.

\subsection{Preliminary}
We first introduce the notations and basic concepts used.
Given a pair of images $I=\{I_1,I_2\}$, 
we denote the motion field as $\vf{f}=\{\vf{f}_{ij}\}$ where $\vf{f}_{ij}=(u_{ij},v_{ij})^T$ for pixels $(i,j)$, $i=1,\dots,n,j=1,\dots,m$. 
In contrast to Wannenwetsch~\etal~\cite{wannenwetsch2017probflow} where $\{\vf{f}_{ij}\}$ are continuous, we treat them as discrete random variables. We call their density functions \emph{match densities}. 
We use $p(\vf{f}|I)$ to denote the joint probability distribution of $\{\vf{f}_{ij}\}$. 
For brevity, we omit the conditional $I$ in the following discussion when there is no ambiguity.
Finally, we introduce a $\times2$ upsampling operator $\varphi$ and an opposite downsampling operator $\varphi^{-1}$.

\subsection{Match Density Decomposition}
\label{sec::hdd}
The main challenge of estimating the full match density is the prohibitive computational cost. Assume we have an image with size $1000\times 1000$ and displacement range $[-50,~50]$. 
The cardinality of $\{\vf{f}_{ij}\}$ would be $10^6$ and the support size 
of each $\vf{f}_{ij}$ could be $10^4$. In this case, the entire distribution volume would have $10$ billion cells, which is intractable to generate.

Our key observation is that the full match density can be decomposed hierarchically into multiple levels of distributions. Fig.~\ref{fig::teaser} provides an intuitive illustration. 
Let us consider multi-scale motion fields $\{\vf{f}^l\}$ $(l=0,\dots,L)$, where higher level $\vf{f}^l$ has half of the resolution of the lower level $\vf{f}^{l+1}$ and $\vf{f}^L$ is identical as $\vf{f}$. 
We introduce a transformation $\vf{g}^l=\vf{f}^l-\varphi(\vf{f}^{l-1})$ to shift the absolute multi-scale motion fields to residual ones. 
We can recover the original motion field $\vf{f}$ from $\{\vf{g}^l\}$ $(l=0,\dots,L)$ via

\begin{equation}
\label{eq::recover}
\vf{f} = \sum_{l=0}^{L} \varphi^{L-l}(\vf{g}^l).
\end{equation}
Naturally, we have the decomposition of $p(\vf{f})$ as

\begin{equation}
\label{eq::decompose}
p(\vf{f})=\sum_{\{\vf{g}^l\}\in \vf{F}} \prod_{l=0}^{L} p(\vf{g}^l|\vf{G}^{l-1}),
\end{equation}
where $\vf{G}^{l}=\{g^s\}_{s=0}^{l}$, and $\vf{F}$ is the set of all possible $\{\vf{g}^l\}$ that satisfies Eq.~\ref{eq::recover}. 
Therefore, we can in turn estimate the decomposed match densities $p(\vf{g}^l|\vf{G}^{l-1})$ 
and recover full match density $p(\vf{f})$ through Eq.~\ref{eq::decompose} afterward. 
The benefit of adopting such decomposition lies in that match density $p(\vf{g}^l|\vf{G}^{l-1})$ actually has quite low variance, \ie probabilities concentrate on a small subset $R'_{\vf{g}^l}$ of the entire support $R_{\vf{g}^l}$. Without loss of much information, we can focus on solving $p(\vf{g}^l|\vf{G}^{l-1})$ with $\vf{g}^s\in R'_{\vf{g}^s}$ for $s=0,\dots,l-1$. 
Consequently, for maximizing the posterior distribution $p(\vf{f})$, we achieve satisfactory approximation through 
maximizing each of the decomposed match densities. 
We will discuss our selection of support subsets in the next section. 

\subsection{Learning Decomposed Match Density}
\label{sec::convert}
Our objective becomes estimating multi-scale decomposed match densities $p(\vf{g}^l|\vf{G}^{l-1})$. 
We propose to learn such information through multiple levels of ConvNets. 
At each level, a ConvNet is designed to estimate the decomposed match density. Note that $\vf{g}^l$ is conditioned on $\vf{G}^{l-1}$, while theoretically we can sample $\vf{g}^s\in\vf{G}^{l-1}$ according to predicted densities at coarser levels.

\begin{figure}[t]
\begin{center}
   \includegraphics[clip, trim=0cm 1.2cm 0cm 0cm, width=1.0\linewidth]{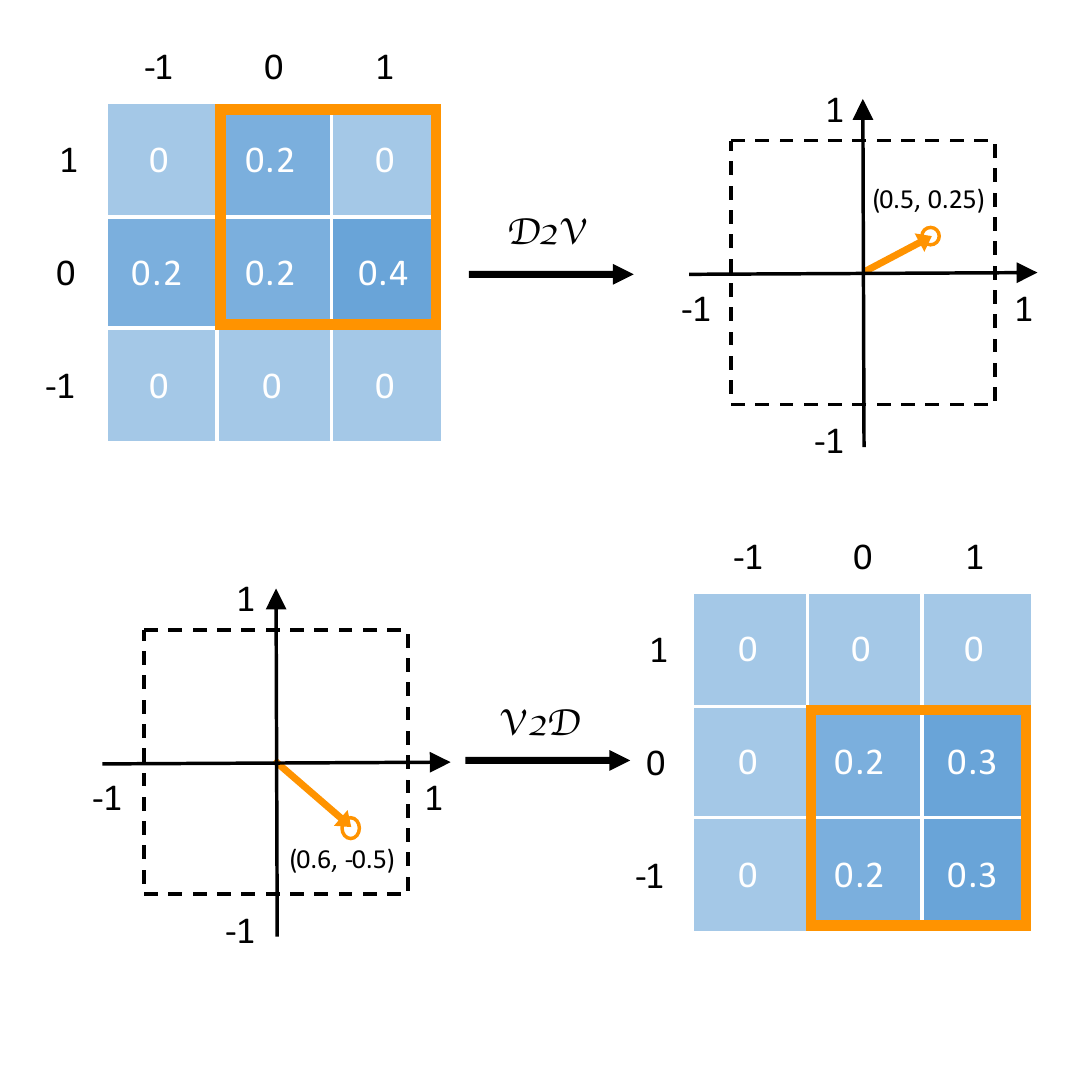}
\end{center}
\vspace{-1ex}
   \caption{Conversion between motion vectors and match densities. The support is taken as $3\times 3$ here for illustration.}
\label{fig::hdd}
\end{figure}

In this section, we first discuss how to transform point estimate into match density, which is adopted for generating our distribution supervision for each level. 
Let us consider a \textit{general} motion vector $\vf{f}_{ij}\in \vf{f}$ and its density function $p(\vf{f}_{ij})$. 
As stated in Sec.~\ref{sec::hdd}, we prefer $p(\vf{f}_{ij})$ to possess a low variance, 
which would greatly reduce the computation cost through our decomposition. 
We observe that real-valued $\vf{f}_{ij}$ uniquely falls into a $2\times 2$ window $W_{ij}$ in the image grid. This inspires us to splat the bilinear weights of $\vf{f}_{ij}$ \wrt coordinates in $W_{ij}$ to $p(\vf{f}_{ij})$. Concretely, for any $\vf{d}\in \mathbb{Z}^2$, we have
\begin{equation}
\label{eq::splat}
\mathbb{P}(\vf{f}_{ij}=\vf{d})=\left\{
\begin{aligned}
0~~~~~~~~~~~~~~~~~~~~~~ & &\vf{d}\notin W_{ij}~ \\
\rho(\vf{f}_{ij}-\tilde{\vf{d}})~~~~~~~~ & &\vf{d}\in W_{ij}, \\
\end{aligned}
\right.
\end{equation}
where $\rho(\cdot)$ means the product of elements in the vector, and $\tilde{\vf{d}}$ is the diagonal opposite coordinate of $\vf{d}$ in $W_{ij}$. We call such conversion as \vtodname (see Fig.~\ref{fig::hdd}), which depicts our assumption for the ground-truth match density.

As seen from Eq.~\ref{eq::splat}, the support of $p(\vf{f}_{ij})$ is indeed $W_{ij}$ which has a maximum size of $4$. 
Ideally, we can sample $\vf{g}^s\in\vf{G}^{l-1}$ in a quadtree fashion during estimating the match density of $\vf{g}^l$. 
However, such computation is still heavy for both training and evaluation. 
For trade-off, we can discard samplings with minor probabilities. 
A trivial practice is always taking $\argmax$ at each level. 
As a substitution, we propose \textit{local expectation} to further reduce the loss of information. 
Specifically, for any \textit{general} match density $p(\vf{f}_{ij})$, we define $W_{ij}^*$ as the $2\times2$ window over which the integral of $p(\vf{f}_{ij})$ maximizes among all candidate windows. 
We only retain the probabilities of $p(\vf{f}_{ij})$ in $W^*_{ij}$ and normalize it into $p^*(\vf{f}_{ij})$. 
The local expectation is defined as $\mathbb{E}[\vf{f}_{ij}]$ \wrt $p^*(\vf{f}_{ij})$. 
In the following, we use expectation to denote local expectation by default.
We call this conversion \dtovname (see Fig.~\ref{fig::hdd}). 
Therefore, at each level, instead of exhaustive sampling we always take the max posterior of $\vf{g}^l$ as $\mathbb{E}[\vf{g}^{l}]$, and we only estimate $p(\vf{g}^l|\vf{G}^{(l-1)*})$ ($p^{l}_{\subfix{res}}$ for short in the following) in each level, where $\vf{G}^{(l-1)*}=\{\mathbb{E}[\vf{g}^s]\}_{s=0}^{l-1}$. 
This enables us to get rid of expensive training and test time sampling.

\begin{figure*}[t]
\centering
\footnotesize
\begin{tabular}{@{}c @{\hskip 0.05in} c @{\hskip 0.05in} c @{\hskip 0.05in} c @{\hskip 0.05in} c @{\hskip 0.05in} c@{}}

\includegraphics[width=0.16\linewidth]{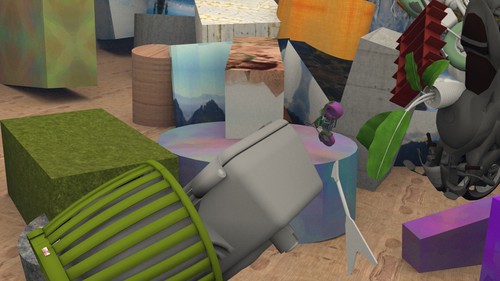}
&
\includegraphics[width=0.16\linewidth]{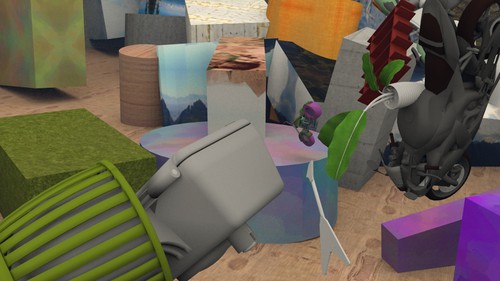}
&
\includegraphics[width=0.16\linewidth]{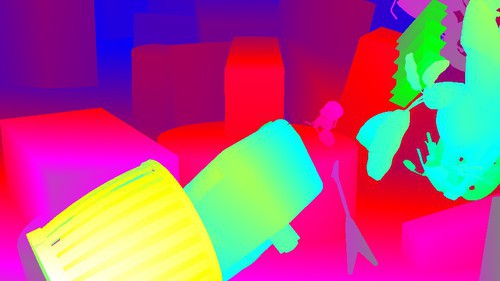}
&
\includegraphics[width=0.16\linewidth]{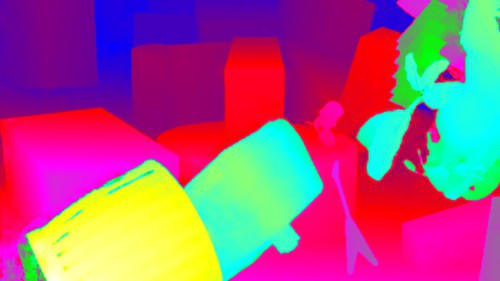}
&
\includegraphics[width=0.16\linewidth]{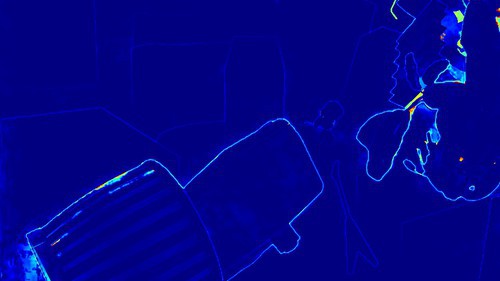}
&
\includegraphics[width=0.16\linewidth]{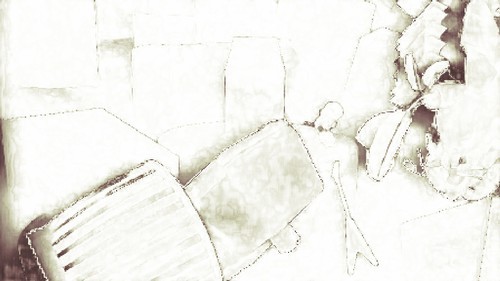}
 \\

\includegraphics[width=0.16\linewidth]{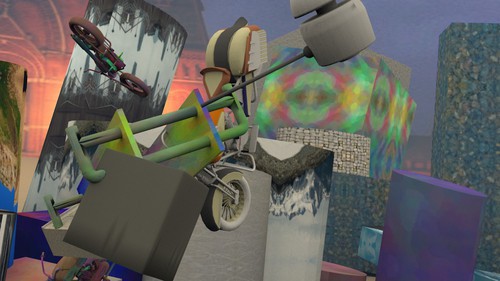}
&
\includegraphics[width=0.16\linewidth]{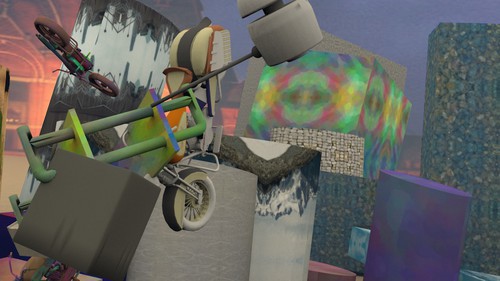}
&
\includegraphics[width=0.16\linewidth]{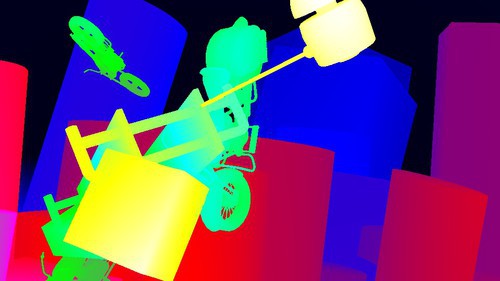}
&
\includegraphics[width=0.16\linewidth]{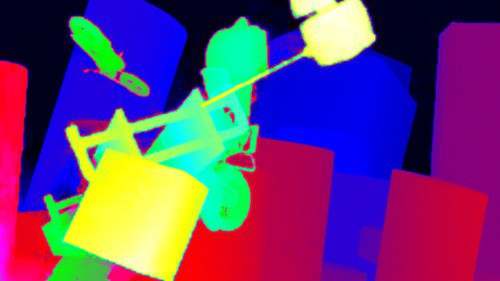}
&
\includegraphics[width=0.16\linewidth]{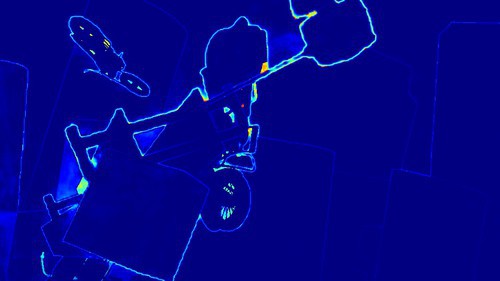}
&
\includegraphics[width=0.16\linewidth]{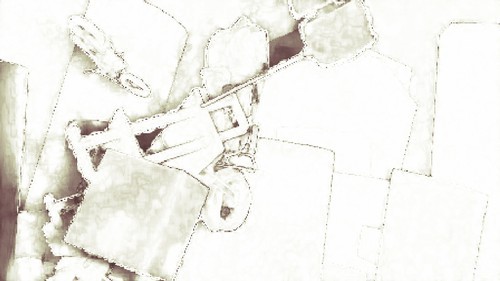}
 \\

\includegraphics[width=0.16\linewidth]{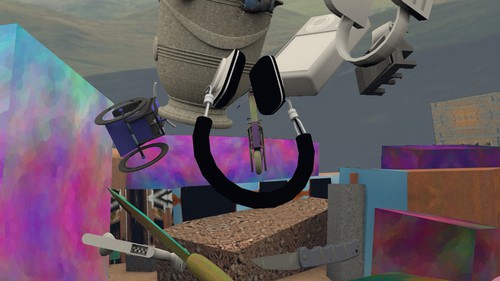}
&
\includegraphics[width=0.16\linewidth]{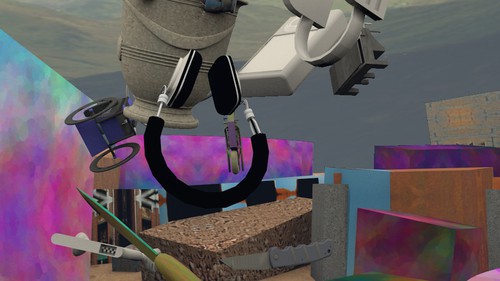}
&
\includegraphics[width=0.16\linewidth]{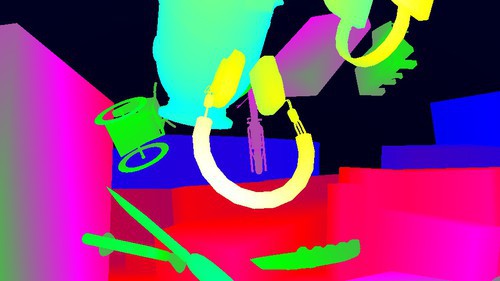}
&
\includegraphics[width=0.16\linewidth]{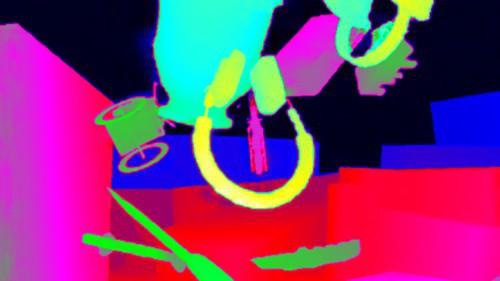}
&
\includegraphics[width=0.16\linewidth]{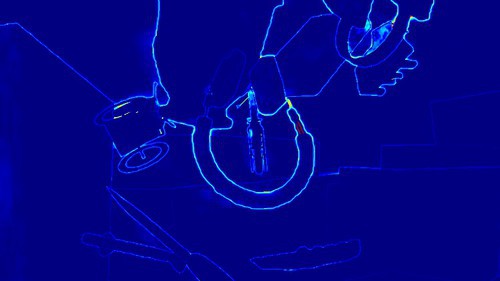}
&
\includegraphics[width=0.16\linewidth]{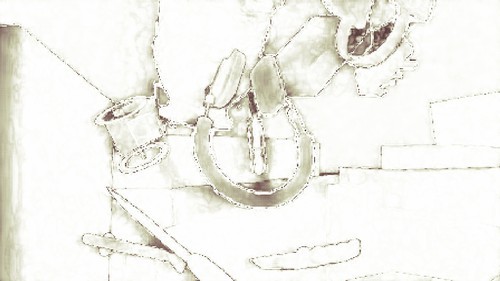}
 \\

Left Image & Right Image & Ground Truth & Prediction & Error Map & Confidence Map
\end{tabular}
\caption{Visualized stereo results on the validation set of FlyingThings3D. Cold colors in the error map denote correct predictions while warm colors mean the contrary. Our network gives accurate results in most regions, while errors tend to occur at boundaries and occlusions.}
\label{fig::things_stereo}
\end{figure*}

\subsection{Network Architecture Design}
\label{sec::archi}
Finally, we discuss the network architecture design for estimating the decomposed match densities. 
We achieve this objective via stacking multiple levels of ConvNets, which we call density decoders $\{D^l\}$. 
$D^l$ infers the match density \plres in its respective level $l$.
A single level of the entire network is illustrated in Fig.~\ref{fig::pipeline}. In the following, we discuss the details of our subnetworks.

The architecture design of our density decoder $D^l$ is motivated by the close relationship between the targeted output $p_{\subfix{res}}^l$ and the similarity information between image pairs, or their embedded representations.
Our match density estimation operates on multi-scale feature embeddings $\{F_1^l,F_2^l\}$, which are extracted via a DLA~\cite{yu2018deep} network over $\{I_1,I_2\}$. 
Affinity information can be obtained through the correlation~\cite{dosovitskiy2015flownet} of feature embeddings between different frames. For performing long-range correlation %
and imposing conditional priors from previous levels, we always warp the feature $F_2^l$ according to $\varphi(\mathbb{E}[\vf{f}^{l-1}])$ before correlation. The cost volume is concatenated with $F_1^l$, $\varphi(\mathbb{E}[\vf{f}^{l-1}])$, and the upsampled \textit{density embedding} $E_{\subfix{up}}^{l-1}$ from the previous level, then fed into our density decoder $D^l$. The decoder $D^l$ %
produces the \textit{density embedding}, from which we obtain the match density via a classifier. Also, we upsample the \textit{density embedding} 
to $E_{\subfix{up}}^l$ and feed it to the next level as density bypass connections. Both of the pyramid feature extractor and the density decoder are jointly trained in an end-to-end manner.

At inference time, we calculate $\mathbb{E}[\vf{g}^l]$ from each predicted \plres and compose them via Eq.~\ref{eq::recover} to produce the point estimate of $\vf{f}$. 
While during training time, we downsample the ground-truth motion field into $\vf{f}_{\subfix{gt}}^l$. The residual motion \wrt $\varphi(\mathbb{E}[\vf{f}^{l-1}])$ 
is converted into $p^l_\subfix{gt}$. 
The entire training loss comes in the form of Kullback$-$Leibler divergence
\begin{equation}
    \label{equa::kl}
\mathcal{L} = \sum_{l}\sum_{\vf{g}\in R_{\vf{g}^l}} p_{\subfix{gt}}^l(\vf{g})(\log p_{\subfix{gt}}^l(\vf{g})-\log p_{\subfix{res}}^l(\vf{g})).
\end{equation}

\section{Experiments}
\begin{figure*}[t]
\centering
\footnotesize
\begin{tabular}{@{}c @{\hskip 0.05in} c @{\hskip 0.05in} c @{\hskip 0.05in} c @{\hskip 0.05in} c @{\hskip 0.05in} c@{}}

\includegraphics[width=0.16\linewidth]{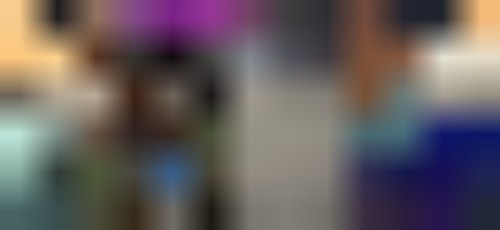}
&
\includegraphics[width=0.16\linewidth]{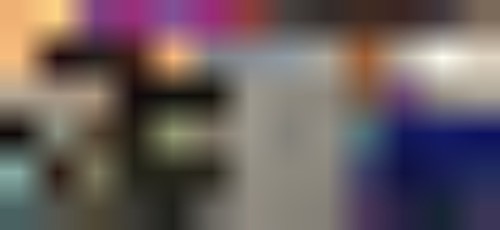}
&
\includegraphics[width=0.16\linewidth]{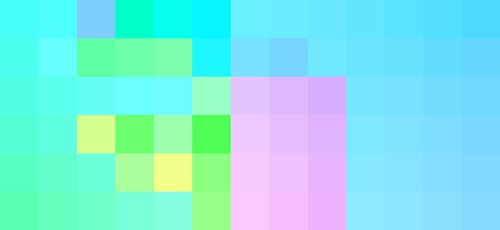}
&
\includegraphics[width=0.16\linewidth]{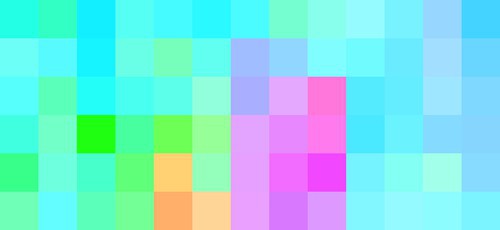}
&
\includegraphics[width=0.16\linewidth]{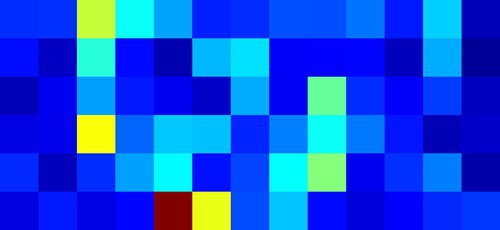}
&
\includegraphics[width=0.16\linewidth]{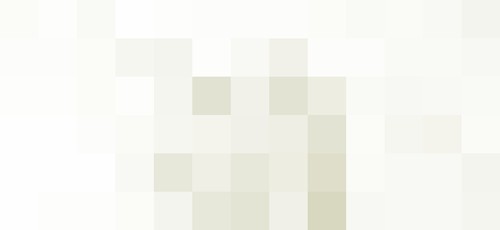}
 \\
\includegraphics[width=0.16\linewidth]{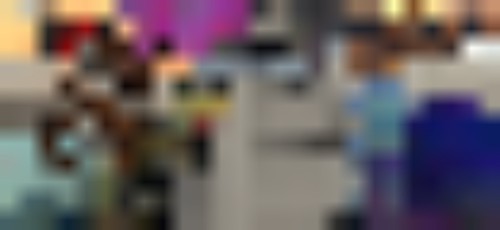}
&
\includegraphics[width=0.16\linewidth]{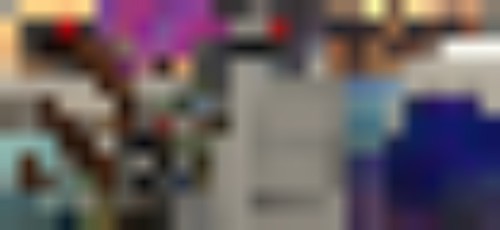}
&
\includegraphics[width=0.16\linewidth]{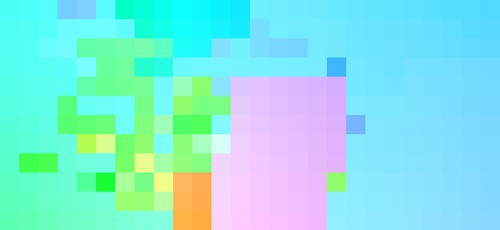}
&
\includegraphics[width=0.16\linewidth]{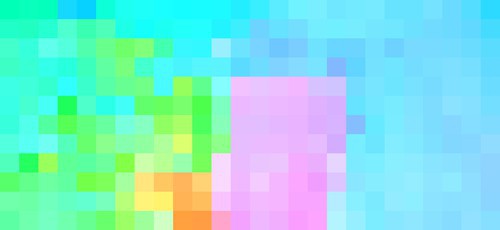}
&
\includegraphics[width=0.16\linewidth]{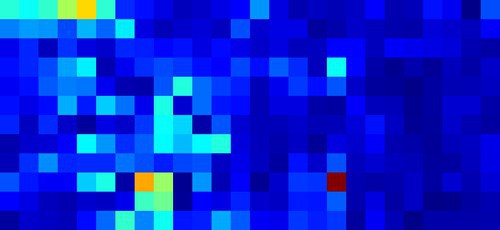}
&
\includegraphics[width=0.16\linewidth]{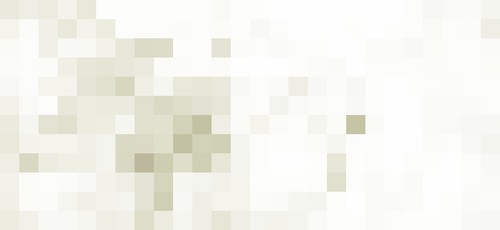}
 \\
\includegraphics[width=0.16\linewidth]{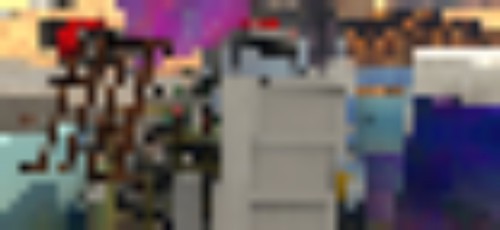}
&
\includegraphics[width=0.16\linewidth]{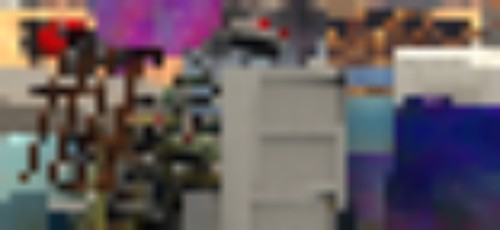}
&
\includegraphics[width=0.16\linewidth]{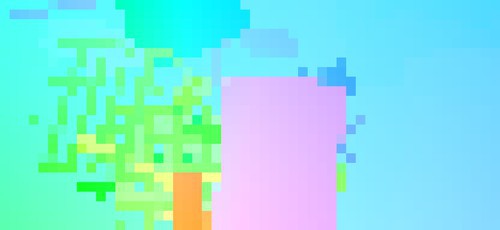}
&
\includegraphics[width=0.16\linewidth]{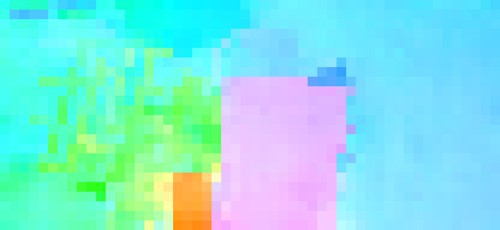}
&
\includegraphics[width=0.16\linewidth]{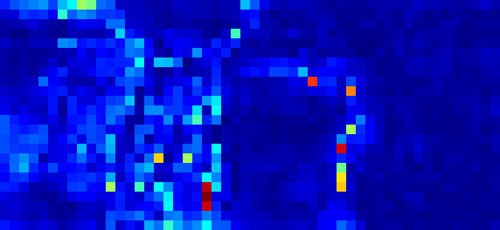}
&
\includegraphics[width=0.16\linewidth]{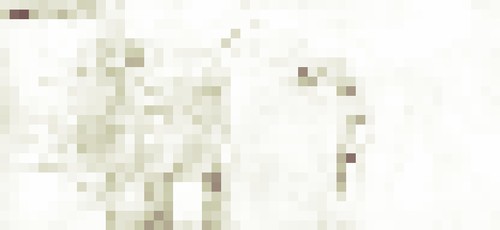}
 \\
\includegraphics[width=0.16\linewidth]{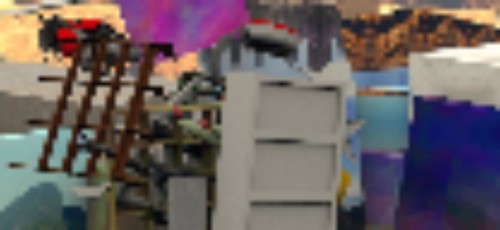}
&
\includegraphics[width=0.16\linewidth]{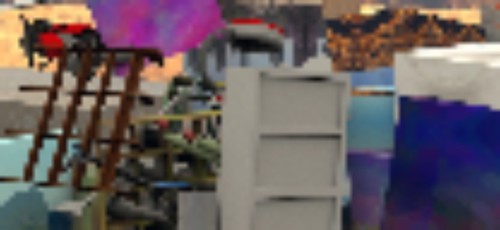}
&
\includegraphics[width=0.16\linewidth]{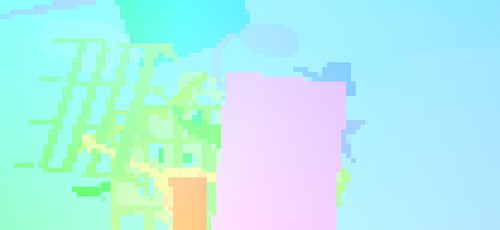}
&
\includegraphics[width=0.16\linewidth]{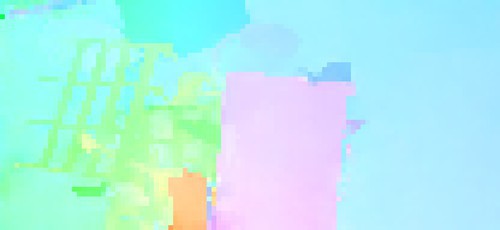}
&
\includegraphics[width=0.16\linewidth]{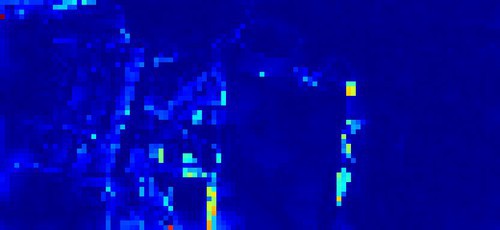}
&
\includegraphics[width=0.16\linewidth]{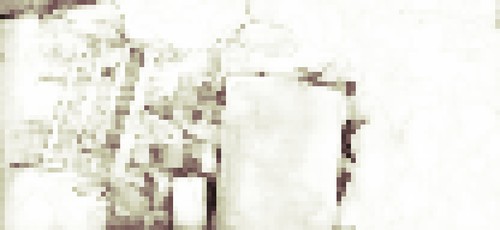}
 \\
\includegraphics[width=0.16\linewidth]{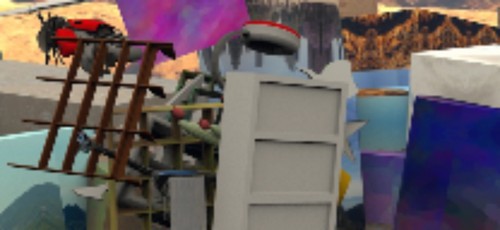}
&
\includegraphics[width=0.16\linewidth]{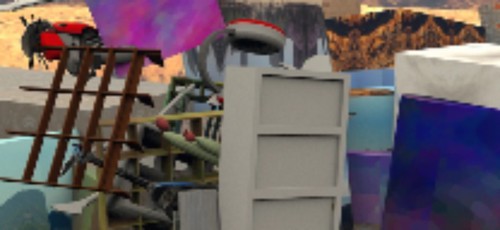}
&
\includegraphics[width=0.16\linewidth]{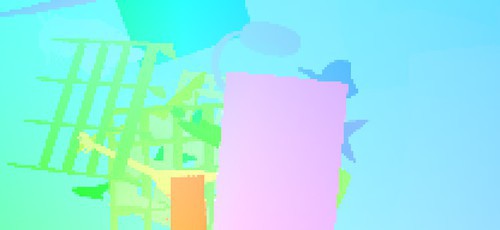}
&
\includegraphics[width=0.16\linewidth]{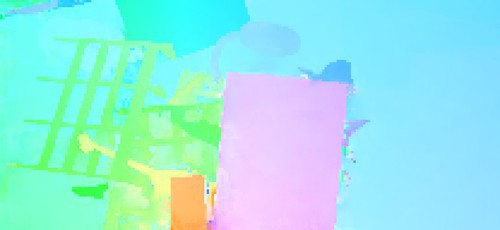}
&
\includegraphics[width=0.16\linewidth]{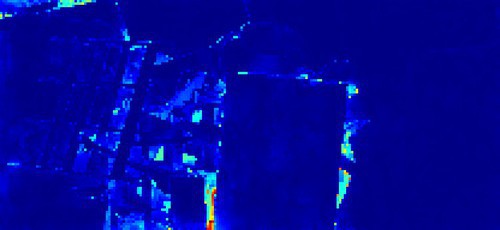}
&
\includegraphics[width=0.16\linewidth]{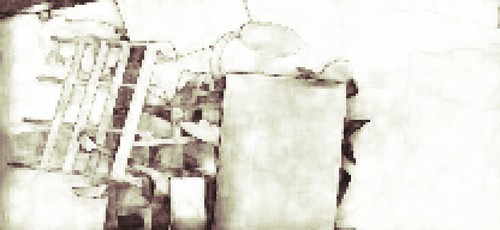}
 \\
Image 1 & Image 2 & Ground Truth & Prediction & Error Map & Confidence Map
\end{tabular}
\caption{Qualitative multi-scale flow result on the validation set of FlyingThings3D dataset. Bilinearly downsampled raw images, coarser level flows, error maps and confidence maps are enlarged via nearest neighbor upsampling for visualization purpose. Our network gives precise predictions in most regions, while occasionally presents confusion in occluded regions and disappearing parts.}
\label{fig::things_flow}
\end{figure*}

\modelname provides hierarchically decomposed match densities. It can be used for different tasks, such as stereo matching and optical flow. The probability of point estimates can be used as uncertainty estimation. 
It is hard to evaluate the quality of the learned distribution directly, but we can investigate its performance being applied to these specific tasks.

\subsection{Implementation Details}
\paragraph{Network Variants.}
We can apply our models to stereo matching and optical flow. 
The networks are called \stereomodelname and \flowmodelname.
The two variants differ slightly: we adopt 1D correlation for \stereomodelname and 2D correlation for \flowmodelname. 
The correlation range is always 4 for both tasks at different levels, which is consistent with the size of match density support. 
Since we treat stereo matching as 1D flow estimation, we clip the positive values in converted point estimates at each level for \stereomodelname. 
The pyramid level is set to $5$ for \flowmodelname and $6$ for \stereomodelname based on experiment results.

\paragraph{Module Details.}
We select DLA-34-Up~\cite{yu2018deep} as our pyramid feature extractor, because it can achieve competitive semantic segmentation accuracy on small datasets with much less computation than the deeper alternatives.
The features at the coarsest level are $\times64$ downsampled.
The density decoder $D^l$ consists of two residual blocks plus one aggregation node~\cite{he2016identity, yu2018deep}, except for the last level when it is fulfilled via a dilated convolutional network~\cite{yu2015multi} as a context module.
We adopt batch normalization~\cite{ioffe2015batch} in all of our models to stabilize the training. 
Predictions are upsampled from the lowest level with highest output resolution to full resolution during evaluation.

\paragraph{Training Details.}
We train our models on $8$ GPUs without synchronized batch normalization.
The weights of pyramid feature extractor are initialized from the ImageNet pretrained model.
The network is optimized by Adam~\cite{kingma2014adam}, where $\beta_1=0.9$, $\beta_2=0.999$. 
For all of our pretraining experiments on synthetic datasets, models are trained for $200$ epochs, and the learning rate is decayed by $0.5$ every $30$ epochs after $70$ epochs for $4$ times in total.
As for data augmentation, besides random cropping, we adopt random resizing and color perturbation~\cite{Meister:2018:UUL} during the fine-tuning stage, and introduce random flipping for optical flow experiments. 
The dense and sparse annotations, as supervision at different scales, are bilinearly downsampled and average pooled from the ground-truth map respectively. 
In this section, unless otherwise stated, confidence maps are obtained through aggregating the probabilities within $W_i^*$ of the last level match density, and uncertainty maps are the opposite.

\begin{table}[tbh]
\small
\centering
\setlength{\tabcolsep}{1pt}
\begin{tabular*}{1.0\linewidth}{r | c  c | c  c  c | p{0.8cm}<{\centering} }
\toprule
& \multicolumn{2}{c|}{KITTI 2012} &  \multicolumn{3}{c|}{KITTI 2015} & Time \\
\midrule
Methods & Out-Noc & Out-All & D1-bg & D1-fg & D1-all & (s) \\
\midrule
SPS-st~\cite{sps} & 3.39 & 4.41 & 3.84 & 12.67 & 5.31 & 2.00 \\
Displets v2~\cite{displet} & 2.37 & 3.09 & 3.00 & 5.56 & 3.43 & 265 \\
MC-CNN-acrt~\cite{mccnn} & 2.43 & 3.63 & 2.89 & 8.88 & 3.88 & 67.0 \\
SGM-Net~\cite{sgmnet} & 2.29 & 3.50 & 2.66 & 8.64 & 3.66 & 67.0 \\
L-ResMatch~\cite{lresmatch} & 2.27 & 3.40 & 2.72 & 6.95 & 3.42 & 48.0\\
GC-Net~\cite{gcnet} & 1.77 & 2.30 & 2.21 & 6.16 & 2.87 & 0.90 \\
EdgeStereo~\cite{edgestereo} & 1.73 & 2.18 & 2.27 & 4.18 & 2.59 & 0.27 \\
PDSNet~\cite{tulyakov2018practical} & 1.92 & 2.53 & 2.29 & 4.05 & 2.58 & 0.50 \\
PSMNet~\cite{chang2018pyramid} & 1.49 & 1.89 & 1.86 & 4.62 & 2.32 & 0.41 \\
SegStereo~\cite{segstereo} & 1.68 & 2.03 & 1.88 & 4.07 & 2.25 & 0.60 \\
\midrule
\stereomodelname~(Ours) & \bf{1.40} & \bf{1.80} & \bf{1.70} & \bf{3.63} & \bf{2.02} & \bf{0.14} \\
\bottomrule
\end{tabular*}
\vspace{0.5ex}
\caption{Stereo matching results on KITTI test set. All of the numbers denote percentages of disparity outliers. The official leaderboard ranks methods according to ``Out-Noc'' for KITTI 2012 and ``D1-all'' for KITTI 2015.}
\vspace{-4ex}
\label{tab:stereo_kitti}
\end{table}

\begin{figure}[t]
\centering
\footnotesize
\begin{tabular}{@{}c @{\hskip 0.025in} c @{\hskip 0.025in} c @{\hskip 0.025in} c@{}}

\includegraphics[width=0.4\linewidth]{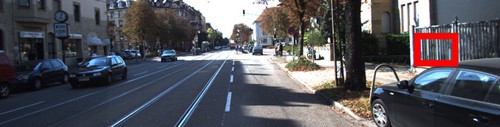}
&
\includegraphics[width=0.4\linewidth]{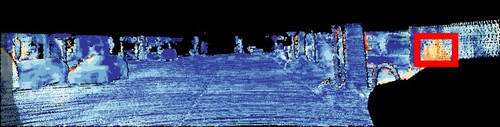}
&
\includegraphics[width=0.137\linewidth]{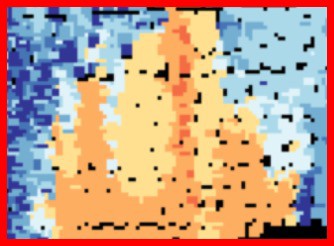}
&
\rotatebox{90}{GCNet}
 \\
\includegraphics[width=0.4\linewidth]{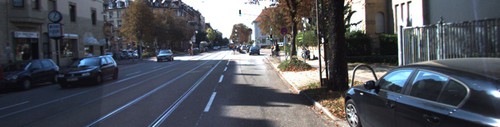}
&
\includegraphics[width=0.4\linewidth]{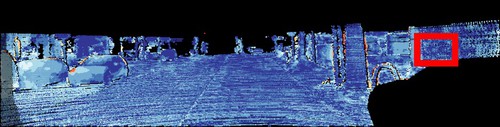}
&
\includegraphics[width=0.137\linewidth]{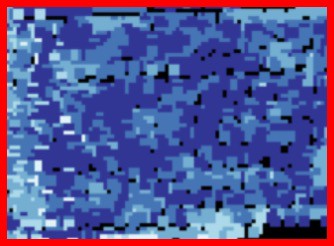}
&
\rotatebox{90}{~~Ours}
 \\  \hline \vspace{-2.3mm} \\
\includegraphics[width=0.4\linewidth]{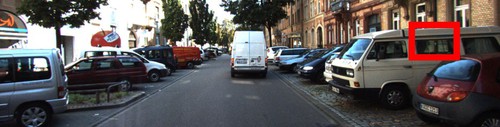}
&
\includegraphics[width=0.4\linewidth]{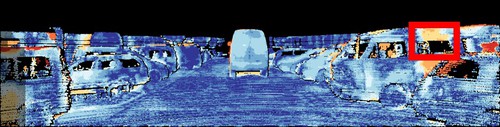}
&
\includegraphics[width=0.137\linewidth]{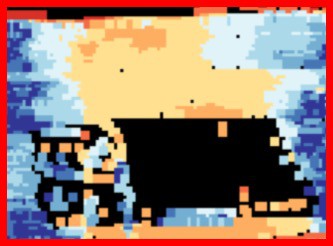}
&
\rotatebox{90}{GCNet}
 \\
 \includegraphics[width=0.4\linewidth]{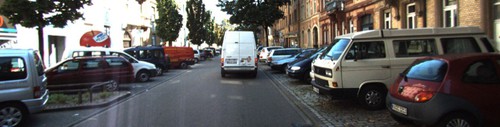}
&
\includegraphics[width=0.4\linewidth]{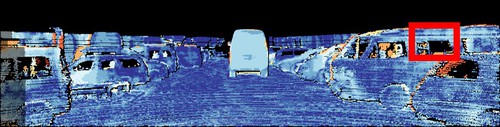}
&
\includegraphics[width=0.137\linewidth]{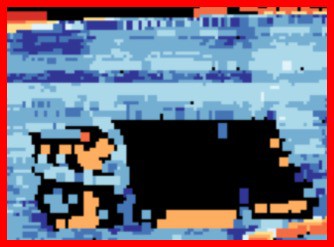}
&
\rotatebox{90}{~~Ours}
 \\  \hline \vspace{-2.3mm} \\
\includegraphics[width=0.4\linewidth]{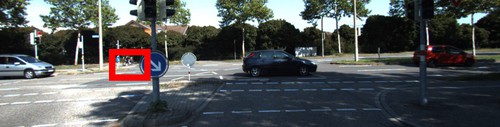}
&
\includegraphics[width=0.4\linewidth]{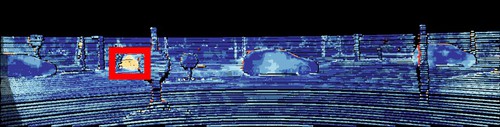}
&
\includegraphics[width=0.137\linewidth]{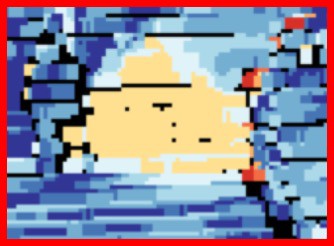}
&
\rotatebox{90}{GCNet}
 \\
\includegraphics[width=0.4\linewidth]{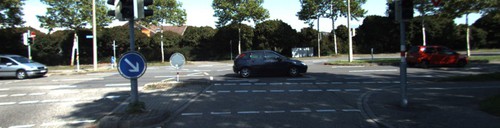}
&
\includegraphics[width=0.4\linewidth]{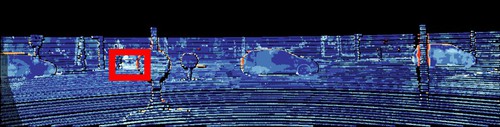}
&
\includegraphics[width=0.137\linewidth]{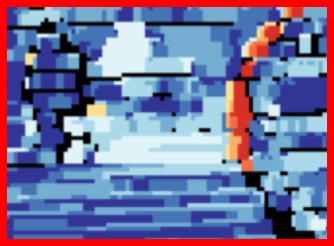}
&
\rotatebox{90}{~~Ours}
 \\
\end{tabular}
\vspace{-1ex}
\caption{Example stereo error maps on KITTI 2015 test set. We contrast our method with GC-Net~\cite{gcnet}. Orange corresponds to erroneous prediction. This figure is best viewed in color.}
\label{fig::kitti_stereo_2015}
\end{figure}

\subsection{Stereo Matching}
To evaluate the performance of our \stereomodelname model, we benchmark our result on the KITTI stereo dataset~\cite{geiger2013vision}. Due to the limited amount of training data in KITTI, we pretrain our model on the FlyingThings3D dataset~\cite{dispnet}.

\minisection{FlyingThings3D.} 
We use the FlyingThings3D dataset as training data. 
Following the training protocol of the original FlowNet2 model~\cite{ilg2017flownet}, we use a subset of the dataset which omits some extremely hard samples.
We train our model with a batch size of $32$ and an initial learning rate of $2\times10^{-4}$. 
The image crop size is $320\times896$.
Qualitative examples, as well as the confidence maps, are shown in Fig.~\ref{fig::things_stereo}. We find low confidence correlates well with prediction errors, which generally occurs at boundaries and occlusions.

\begin{table}[tbh]
\small
\centering
\setlength{\tabcolsep}{5pt}
\begin{tabular*}{1.0\linewidth}{r | c  c | c  c | c}
\toprule
& \multicolumn{2}{c|}{Training} &  \multicolumn{2}{c|}{Test} & Time \\
\midrule
Methods & Clean & Final & Clean & Final & (s) \\
\midrule
PatchBatch~\cite{patchbatch} & - & - & 5.79 & 6.78 & 50.0 \\
EpicFlow~\cite{epic} & - & - & 4.12 & 6.29 & 15.0 \\
CPM-flow~\cite{cpm} & - & - & 3.56 & 5.96 & 4.30 \\
FullFlow~\cite{full} & - & 3.60 & \bf{2.71} & 5.90 & 240 \\
FlowFields~\cite{bailer2015flow} & - & - & 3.75 & 5.81 & 28.0 \\
MRFlow~\cite{mrflow} & 1.83 & 3.59 & 2.53 & 5.38 & 480 \\
FlowFieldsCNN~\cite{flowfieldscnn} & - & - & 3.78 & 5.36 & 23.0 \\
DCFlow~\cite{xu2017accurate} & - & - & 3.54 & 5.12 & 8.60 \\
SpyNet-ft~\cite{ranjan2017optical} & (3.17) & (4.32) & 6.64 & 8.36 & 0.16 \\
FlowNet2~\cite{ilg2017flownet} & 2.02 & 3.14 & 3.96 & 6.02 & 0.12 \\
FlowNet2-ft~\cite{ilg2017flownet} & \bf{(1.45)} & (2.01) & 4.16 & 5.74 & 0.12 \\
LiteFlowNet~\cite{hui2018liteflownet} & 2.52 & 4.05 & - & - & 0.09 \\
LiteFlowNet-ft~\cite{hui2018liteflownet} & (1.64) & (2.23) & 4.86 & 6.09 & 0.09 \\
PWC-Net~\cite{sun2018pwc} & 2.55 & 3.93 & - & - & \bf{0.03} \\
PWC-Net-ft~\cite{sun2018pwc} & (2.02) & (2.08) & 4.39 & 5.04 & \bf{0.03} \\
\midrule
\flowmodelname (Ours) & 3.84 & 8.77 & - & - & 0.08 \\
\flowmodelname-ft (Ours) & (1.70) & \bf{(1.17)} & 4.79 & \bf{4.67} & 0.08 \\
\bottomrule
\end{tabular*}
\vspace{0.5ex}
\caption{Average EPE results on MPI Sintel dataset. ``-ft'' means finetuning on the Sintel training set and numbers in the parenthesis are results on data the method has been trained on.}
\label{tab:flow_sintel}
\end{table}

\minisection{KITTI.} During fine-tuning stage, we leverage the available $394$ image pairs from KITTI 2012 \& 2015 as training data. 
Training is performed for $2000$ epochs, with batch size $16$ and image crop size $320\times896$. 
The initial learning rate is $1\times10^{-5}$ and decayed by $0.5$ at the $1000\subfix{th}$ and the $1500\subfix{th}$ epoch. 

As shown in Tab.~\ref{tab:stereo_kitti}, our method achieves the lowest percentages of disparity outliers in both non-occluded (Out-Noc) and total regions (Out-All), background (D1-bg) and foreground regions (D1-fg), among all of the competitive baselines on both datasets. 
We also hold the lowest inference time for processing a standard KITTI stereo pair. Note that we do not leverage the entire Scene Flow dataset~\cite{dispnet} for training as~\cite{chang2018pyramid, gcnet}, nor do we utilize additional semantic or edge cues as in~\cite{edgestereo, segstereo}.
Qualitative comparisons are shown in Fig.~\ref{fig::kitti_stereo_2015}. Our method exhibits better performance in regions with complex and ambiguous textures. This indicates the effectiveness of hierarchical match density learning based on pyramid feature representations, which exhibits robustness to local noise. 

\subsection{Optical Flow}
We pretrain our \flowmodelname on synthetic data from FlyingChairs~\cite{dosovitskiy2015flownet} and FlyingThings3D~\cite{ilg2017flownet}, then investigate the effectiveness of our model on established optical flow benchmarks including MPI Sintel~\cite{sintel} and KITTI~\cite{geiger2013vision}.

\minisection{FlyingChairs.} We train our network on FlyingChairs with batch size $64$ and initial learning rate $4\times 10^{-4}$. Images are randomly resized and cropped to $384\times512$ patches. We find larger crop size can improve the network performance.

\begin{table}[tbh]
\small
\centering
\setlength{\tabcolsep}{1pt}
\begin{tabular*}{1.0\linewidth}{r | c  c  c | c  c  c}
\toprule
& \multicolumn{3}{c|}{KITTI 2012} &  \multicolumn{3}{c}{KITTI 2015} \\
\midrule
Methods & AEPE & AEPE & F1-Noc & AEPE & F1-all & F1-all\\
 & \emph{train} & \emph{test} & \emph{test} & \emph{train} & \emph{train} & \emph{test} \\
\midrule
EpicFlow~\cite{epic} & - & 3.8 & 7.88\%  &  - & - & 26.29\% \\
FullFlow~\cite{full} & - & - & - & - & - & 23.37\% \\
PatchBatch~\cite{patchbatch} & - & 3.3 & 5.29\% & - & - & 21.07\% \\
FlowFields~\cite{bailer2015flow} & - & - & - & - & - & 19.80\% \\
DCFlow~\cite{xu2017accurate} & - & - & - & - & 15.09\% & 14.83\% \\
MirrorFlow~\cite{mirror} & - & 2.6 & 4.38\% & - & 9.93\% & 10.29\% \\
PRSM~\cite{prsm} & - & \bf{1.0} & 2.46\% & - & - & 6.68\% \\
SpyNet-ft~\cite{ranjan2017optical} & (4.13) & 4.7 & 12.31\% & - & - & 35.07\% \\
FlowNet2~\cite{ilg2017flownet} & 4.09 & - & - & 10.06 & 30.37\% & - \\
FlowNet2-ft~\cite{ilg2017flownet} & (1.28) & 1.8 & 4.82\% & (2.30) & (8.61\%) & 10.41\% \\
LiteFlowNet~\cite{hui2018liteflownet} & 4.25 & - & - & 10.46 & 29.30\% & -\\
LiteFlowNet~\cite{hui2018liteflownet} & (1.26) & 1.7 & - & (2.16) & (8.16\%) & 10.24\% \\
PWC-Net~\cite{sun2018pwc} & 4.14 & - & - & 10.35 & 33.67\% & - \\
PWC-Net-ft~\cite{sun2018pwc} & (1.45) & 1.7 & 4.22\% & (2.16) & (9.80\%) & 9.60\% \\
\midrule
\flowmodelname (Ours) & 4.65 & - & - & 13.17 & 23.99\% & - \\
\flowmodelname-ft (Ours) & \bf{(0.81)} & 1.4 & \bf{2.26\%} & \bf{(1.31)} & \bf{(4.10\%)} & \bf{6.55\%} \\
\bottomrule
\end{tabular*}
\vspace{0.5ex}
\caption{Optical flow results on KITTI dataset. ``-ft'' means fine-tuning on the KITTI training set. Numbers in parenthesis are results on data the network has been trained on.}
\label{tab:flow_kitti}
\end{table}

\begin{figure}[t]
\centering
\footnotesize
\begin{tabular}{@{}c @{\hskip 0.025in} c @{\hskip 0.025in} c @{\hskip 0.025in} c@{}}

\includegraphics[width=0.4\linewidth]{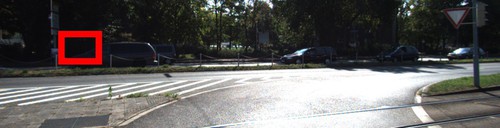}
&
\includegraphics[width=0.4\linewidth]{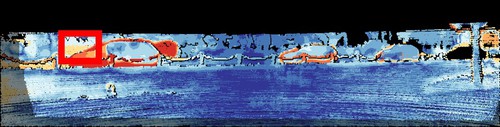}
&
\includegraphics[width=0.139\linewidth]{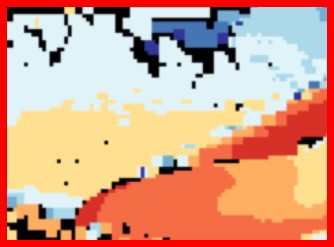}
&
\rotatebox{90}{~~PWC}
 \\
\includegraphics[width=0.4\linewidth]{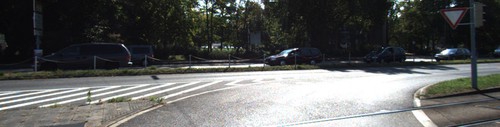}
&
\includegraphics[width=0.4\linewidth]{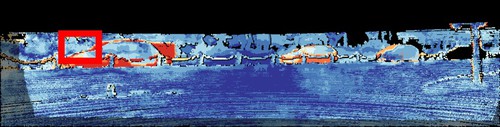}
&
\includegraphics[width=0.139\linewidth]{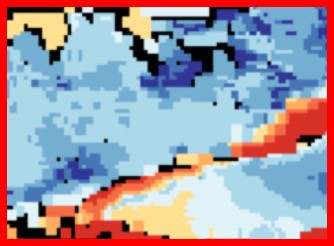}
&
\rotatebox{90}{~~Ours}
 \\ \hline \vspace{-2.3mm} \\ 
\includegraphics[width=0.4\linewidth]{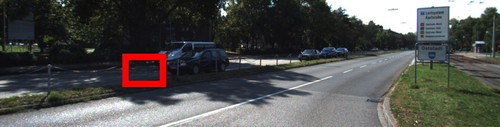}
&
\includegraphics[width=0.4\linewidth]{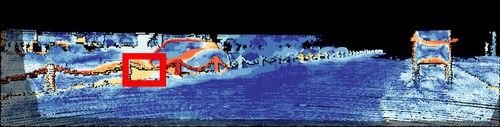}
&
\includegraphics[width=0.139\linewidth]{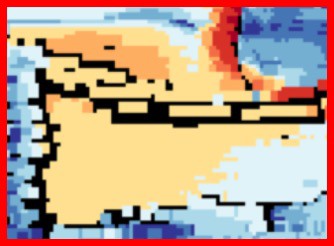}
&
\rotatebox{90}{~~PWC}
 \\
 \includegraphics[width=0.4\linewidth]{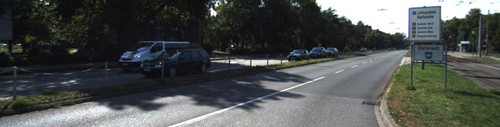}
&
\includegraphics[width=0.4\linewidth]{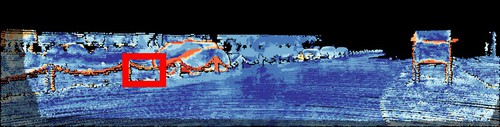}
&
\includegraphics[width=0.139\linewidth]{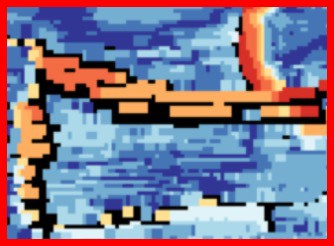}
&
\rotatebox{90}{~~Ours}
 \\  \hline \vspace{-2.3mm} \\
\includegraphics[width=0.4\linewidth]{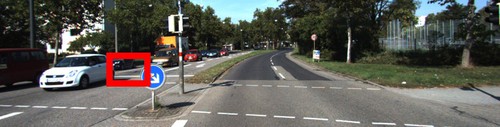}
&
\includegraphics[width=0.4\linewidth]{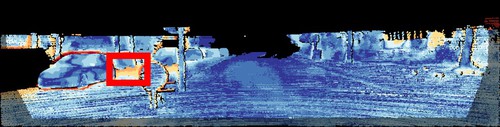}
&
\includegraphics[width=0.139\linewidth]{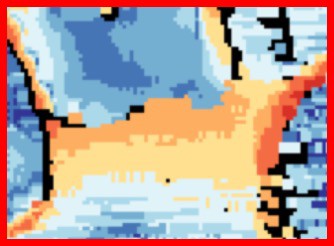}
&
\rotatebox{90}{~~PWC}
 \\
\includegraphics[width=0.4\linewidth]{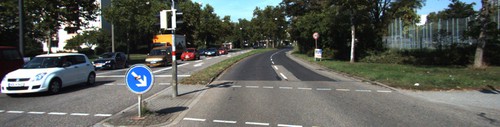}
&
\includegraphics[width=0.4\linewidth]{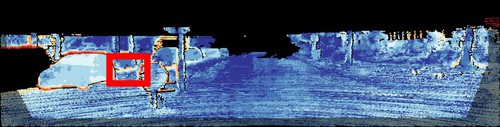}
&
\includegraphics[width=0.139\linewidth]{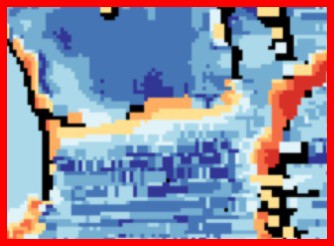}
&
\rotatebox{90}{~~Ours}
 \\
\end{tabular}
\vspace{-1ex}
\caption{Example flow error maps on KITTI 2015 test set. We compare our method with PWC-Net~\cite{sun2018pwc}. Orange corresponds to erroneous prediction. This figure is best viewed in color.} 
\label{fig::kitti_flow_2015}
\end{figure}

\minisection{FlyingThings3D.}
We further fine-tune the model on the FlyingThings3D data, the same subset in our stereo matching experiments, with batch size $32$, learning rate $4\times 10^{-5}$ and image crop size $384\times832$.
We visualize examples of multi-scale predictions in Fig.~\ref{fig::things_flow}. The results indicate that our model is able to progressively refine the prediction from coarse to fine scales. Though we adopt the discrete distribution, our model can still capture very detailed displacements.

\minisection{MPI Sintel.} Finally, we fine-tune our model on MPI Sintel~\cite{sintel} for $1200$ epochs with batch size $32$ and image crop size $384\times768$. 
The initial learning rate is $2\times 10^{-5}$ and decayed by $0.5$ at the $600\subfix{th}$ and the $900\subfix{th}$ epoch.  
Though the dataset provides training data of different subsets (\emph{clean} \& \emph{final} passes), we only adopt the \emph{final} pass as training data rendered with motion blur, defocus blur, and atmospheric effects. %
As shown in Tab.~\ref{tab:flow_sintel}, we can obtain the lowest average EPE in the \emph{final} pass, and compelling results on the \emph{clean} pass, though our model does not see the \emph{clean} pass data during training.
In the model generalization experiment, 
our pretrained \flowmodelname estimates the flow accurately near the occlusion boundary, resulting in the lowest outlier percentage on KITTI (see the ``\flowmodelname (Ours)'' entry in Tab.~\ref{tab:flow_kitti}). 
The metric of EPE emphasizes large motion error. This influence makes our pretrained \flowmodelname achieve higher EPE on MPI Sintel (see the ``\flowmodelname (Ours)'' entry in Tab.~\ref{tab:flow_sintel}). 

\minisection{KITTI.}
Alternatively, we can finetune our pretrained model on KITTI dataset. 
We follow the configurations of our stereo experiment. %
Tab.~\ref{tab:flow_kitti} summarize the results. 
We can obtain the lowest F1-Noc on KITTI 2012 test set and the lowest F1-all on KITTI 2015 test set. 
At the time of writing, \flowmodelname outperforms all two-frame optical flow methods by large margins on both KITTI 2012 \& 2015. 
It even surpasses some competitive scene flow methods such as PRSM~\cite{prsm}, which use additional stereo data. 
This reveals the suitability of our probabilistic method in challenging real-world cases. We show qualitative comparisons against PWC-Net in Fig.~\ref{fig::kitti_flow_2015}. Our method appear to have advantages in estimating many thin structures.

\begin{figure}[t]
\centering
\small
\begin{tabular}{@{}c @{\hskip 0.04in} c@{}}
\multicolumn{2}{c}{\footnotesize Input Image}  \\
\includegraphics[width=0.49\linewidth]{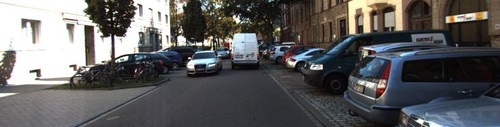}
&
\includegraphics[width=0.49\linewidth]{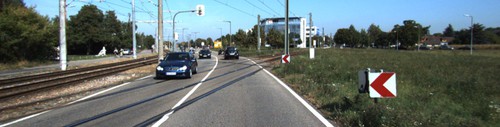}
 \\
\multicolumn{2}{c}{\footnotesize Confidence Map} \\
\includegraphics[width=0.49\linewidth]{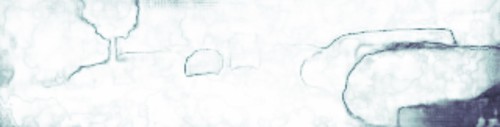}
&
\includegraphics[width=0.49\linewidth]{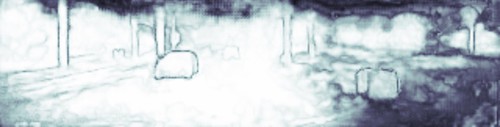}
 \\
\multicolumn{2}{c}{\footnotesize Error Map} \\
\includegraphics[width=0.49\linewidth]{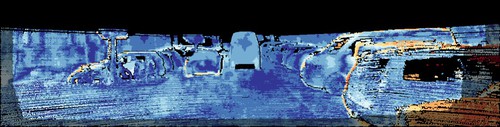}
&
\includegraphics[width=0.49\linewidth]{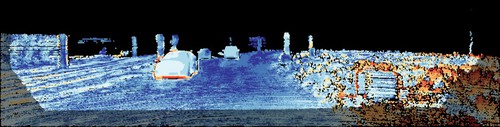}
\end{tabular}
\vspace{-1ex}
\caption{Example confidence maps of our predictions and error maps \wrt ground-truth. In confidence maps, white colors mean confident predictions while dark colors denote uncertain ones. In the error maps, warmer colors indicate inaccurate predictions.}
\label{fig::uncertain}
\end{figure}

\subsection{Uncertainty Estimation}
We also conduct quantitative analysis of uncertainty estimation. 
We compute the log likelihoods of our network predictions and compare \flowmodelname with probabilistic flow networks~\cite{gast2018lightweight}. 
FlowNetDropOut uses variational Gaussian dropout layers~\cite{kingma2015variational}. While FlowNetProbOut replaces deterministic outputs with probabilistic output layers. FlowNetADF propagates uncertainty through the entire network using ADF. 
During the evaluation, we recover the full match density through composing the multi-scale match densities. 
This can be achieved through iteratively sampling from coarse to fine, and we assume a discrete non-uniform distribution for sampling outside $W^*_i$ (see Sec.~\ref{sec::convert}). 
As shown in Tab.~\ref{tab:flow_uncertainty}, \flowmodelname achieves the best %
average log likelihoods against all of the baselines.

\begin{table}[tbh]
\small
\centering
\setlength{\tabcolsep}{5pt}
\begin{tabular*}{1.0\linewidth}{r | c c c }
\toprule
Methods & Sintel clean & Sintel final & Chairs \\
\midrule
FlowNetDropOut~\cite{gast2018lightweight} & -7.106 & -10.820 & -6.176 \\
FlowNetProbOut~\cite{gast2018lightweight} & -6.888 & -7.621 & -3.591 \\
FlowNetADF~\cite{gast2018lightweight} & -3.878 & -4.186 & -3.348 \\
\midrule
HD$^3$F(Ours) & \bf{-1.487} & \bf{-1.821} & \bf{-0.872} \\
\bottomrule
\end{tabular*}
\caption{Average log likelihoods of probabilistic flow methods on MPI Sintel training set and FlyingChairs test set.}
\vspace{-1ex}
\label{tab:flow_uncertainty}
\end{table}

\begin{table}[tbh]
\small
\centering
\setlength{\tabcolsep}{8.5pt}
\begin{tabular*}{1.0\linewidth}{c | c | c  c | c  c}
\toprule
\multirow{2}{*}{Classes} & \multirow{2}{*}{Methods} & \multicolumn{2}{c|}{Noc} &  \multicolumn{2}{c}{All} \\
 & & IoU & Acc & IoU & Acc\\
\midrule
\multirow{2}{*}{Outlier} & Consistency & 17.5 & \bf{64.9} & 23.3 & \bf{81.9} \\
& Ours & \bf{37.6} & 57.8 & \bf{44.1} & 76.4 \\
\midrule
\multirow{2}{*}{Inlier}& Consistency & 84.2 & 85.8 & 75.6 & 76.9 \\
& Ours & \bf{96.1} & \bf{97.8} & \bf{91.8} & \bf{93.7} \\
\midrule
\multirow{2}{*}{Mean}& Consistency & 50.9 & 75.4 & 49.5 & 79.4 \\
& Ours & \bf{66.9} & \bf{77.8} & \bf{68.0} & \bf{85.1} \\
\bottomrule
\end{tabular*}
\vspace{0.5ex}
\caption{Classification result of inlier and outlier predictions on KITTI 2015 training set. Noc denotes evaluation only in the non-occluded area, while All denotes evaluation in the overall region.}
\vspace{-1ex}
\label{tab:uncer_error}
\end{table}

Furthermore, we measure the reliability of network prediction based on uncertainty. We treat predictions with uncertainty greater than a certain threshold ($\sigma=0.3$) as outliers and compare such criterion with the forward-backward consistency check~\cite{yin2018geonet} which is popularly adopted for point estimate. Both methods use the same \flowmodelname model for inference. As shown in Tab.~\ref{tab:uncer_error}, our uncertainty estimation gives the highest mean IoU and mean accuracy in both non-occluded and overall regions. 
Fig.~\ref{fig::uncertain} presents visualization of the confidence and error maps.
We can observe the positive correlation between our estimated uncertainty and prediction error.

\section{Conclusion}
We proposed Hierarchical Discrete Distribution Decomposition (HD$^3$) for estimating the match density. 
Our approach decomposed the match density into multiple scales and learned the decomposed match densities in an end-to-end manner. 
The predicted match densities can be converted into point estimates, while providing model-inherent uncertainty measures at the same time. Our experiments demonstrated the advantages of our method on established benchmarks.

In the future, we hope to integrate more information into our framework %
such as the pixel assignment probabilities from segmentation. %
Currently, we do not consider relationships between match densities of adjacent pixels, but this may help remove match uncertainty in challenging cases.

\paragraph{Acknowledgments} This work was supported by Berkeley AI Research and Berkeley DeepDrive.

{\small
\bibliographystyle{ieee}
\bibliography{references}
}

\end{document}


\title{Hierarchical Discrete Distribution Decomposition for Match Density Estimation \\
\textcolor{red}{\large Supplementary Material}}

\author{First Author\\
Institution1\\
Institution1 address\\
{\tt\small firstauthor@i1.org}
\and
Second Author\\
Institution2\\
First line of institution2 address\\
{\tt\small secondauthor@i2.org}
}

\maketitle

\section{Quantitative Analysis of Uncertainty}
Here we investigate how to distinguish correct/wrong predictions based on our uncertainty estimation, and compare it with widely-adopted method for vectorized flow representation. For simplicity, we take optical flow setting on KITTI 2015 dataset for illustration. 

Naturally, the per-pixel classification of predictions into correct/wrong ones can be treated as a binary segmentation task. For our probabilistic model, predictions with high uncertainty are more likely to be incorrect.
Thus we simply set a threshold $\sigma$ ($0<\sigma<1$), and classify predictions with uncertainty larger (or less) than $\sigma$ as outliers (or inliers accordingly). In our experiment, we set $\sigma$ to be $0.3$. 
For comparison, we adopt the traditional forward-backward consistency check for detecting erroneous results. Predictions where the forward and backward flows contradict seriously are treated as outliers. The threshold follows~\cite{yin2018geonet}. Both methods adopt the same HD$^3$F\xspace model for inference.

\begin{table}[tbh]
\small
\centering
\setlength{\tabcolsep}{8.5pt}
\begin{tabular*}{1.0\linewidth}{c | c | c  c | c  c}
\toprule
&  & \multicolumn{2}{c|}{Noc} &  \multicolumn{2}{c}{All} \\
Classes & Methods & IoU & Acc & IoU & Acc\\
\midrule
\multirow{2}{*}{Outlier} & Consistency & 17.5 & \bf{64.9} & 23.3 & \bf{81.9} \\
& Ours & \bf{37.6} & 57.8 & \bf{44.1} & 76.4 \\
\midrule
\multirow{2}{*}{Inlier}& Consistency & 84.2 & 85.8 & 75.6 & 76.9 \\
& Ours & \bf{96.1} & \bf{97.8} & \bf{91.8} & \bf{93.7} \\
\midrule
\multirow{2}{*}{Mean}& Consistency & 50.9 & 75.4 & 49.5 & 79.4 \\
& Ours & \bf{66.9} & \bf{77.8} & \bf{68.0} & \bf{85.1} \\
\bottomrule
\end{tabular*}
\vspace{0.5ex}
\caption{Inlier/outlier prediction classification result on KITTI 2015 training set. Noc denotes evaluation only in the non-occluded area, while All denotes evaluation in overall region.}
\label{tab:uncer_error}
\end{table}

As can be seen in Tab.~\ref{tab:uncer_error}, criteria provided by our uncertainty map gives the highest mean IoU and mean Accuracy in both non-occluded and overall regions. It manifests the positive correlation between our uncertainty estimation and prediction error distribution. Our method requires less computation (consistency check needs computing both forward and backward flows) while outperforms hand-crafted heuristic method.

\section{Quantitative Region Propagation Result}
Now we conduct quantitative analysis of our probabilistic model for region propagation. Due to the limitation of available groundtruth optical flow and video semantic segmentation, we train a state-of-the-art segmentation network DLA-102~\cite{yu2018deep} on KITTI. Then we generate video semantic segmentation predictions on KITTI as pseudo labels. Based on these generated annotations, we aim to propagate the segmentation of first frame to the following $T$ frames only relying on correspondence information. We achieve this via forward splatting~\cite{lsiTulsiani18} guided by probabilistic or vectorized flow. The classification probability is propagated to following frames sequentially, and we always select the classes with highest probabilities as predictions of current frame. We evaluate the performance in terms of semantic segmentation metric and compare our method with~\cite{sun2018pwc} in Tab.~\ref{tab::region_prop}. Our method achieves better mean IoU and mean Accuracy scores against representative flow regression work~\cite{sun2018pwc}.

\begin{table}[tbh]
\small
\centering
\setlength{\tabcolsep}{11pt}
\begin{tabular*}{0.8\linewidth}{r | c | c}
\toprule
Method & mean IoU & mean Acc\\
\midrule
PWCNet & 38.3\% & 60.5\%  \\
Ours HD$^3$F\xspace & \bf{39.7\%} & \bf{63.6\%}\\
\bottomrule
\end{tabular*}
\vspace{0.5ex}
\caption{Region propagation result on KITTI multiview extension dataset. A total of 150 independent sequences with sequence length of 10 are selected for evaluation.}
\label{tab::region_prop}
\end{table}

\section{Ablation Study with Feature Extractor}
It is well known that deeper representations can give promisingly better performance in a wide variety of computer vision tasks. While off-the-shelf models~\cite{ilg2017flownet, sun2018pwc} in correspondence learning still adopt VGG-like encoders. Here we study how deeper representations can improve the performance of our probabilistic model. We conduct experiment on KITTI stereo dataset with our HD$^3$S\xspace model.

For comparison, we adopt DLA-34 and DLA-34-up~\cite{yu2018deep} as two options for our pyramid feature constructor. The latter one appends
a progressively deeper and higher resolution decoder via iterative deep aggregation. As can be seen in Tab.~\ref{tab::stereo_ablation}, deeper hierarchical representations give stable improvement on both KITTI 2012 \& 2015 datasets.

\begin{table}[tbh]
\small
\centering
\setlength{\tabcolsep}{6pt}
\begin{tabular*}{1.0\linewidth}{r | c c | c  c  c}
\toprule
& \multicolumn{2}{c|}{KITTI 2012} &  \multicolumn{3}{c}{KITTI 2015} \\
\midrule
Encoder & Out-Noc & Out-All & D1-bg & D1-fg & D1-all\\
\midrule
DLA & 1.83\% & 2.43\% & 2.12\% & \bf{5.40\%} & 2.67\% \\
DLA-Up & \bf{1.68\%} & \bf{2.25\%} & \bf{2.06\%} & 5.47\% & \bf{2.63\%} \\
\bottomrule
\end{tabular*}
\vspace{0.5ex}
\caption{Stereo matching result on KITTI official test set with different feature encoder architecture. All of the numbers denote percentage of pixels with errors of more than three pixels or 5\% of disparity magnitude.}
\label{tab::stereo_ablation}
\end{table}

{\small
\bibliographystyle{ieee}
\bibliography{egbib}
}
\clearpage